\documentclass[conference]{IEEEtran}
\IEEEoverridecommandlockouts

\usepackage{cite}
\usepackage{amsmath,amssymb,amsfonts}
\usepackage{algorithmic}
\usepackage{graphicx}
\usepackage{textcomp}
\usepackage{xcolor}
\usepackage{algorithm}
\usepackage{booktabs}
\usepackage{subcaption}
\usepackage{multirow}

\def\BibTeX{{\rm B\kern-.05em{\sc i\kern-.025em b}\kern-.08em
    T\kern-.1667em\lower.7ex\hbox{E}\kern-.125emX}}
\begin{document}

\title{Two Heads are Better than One: Robust Learning Meets Multi-branch Models}
\author{
    Zongyuan Zhang\textsuperscript{\rm \dag},
    Qingwen Bu\textsuperscript{\rm \ddag},
    Tianyang Duan\textsuperscript{\rm \dag},
    Zheng Lin\textsuperscript{\rm \S}, 
    Yuhao Qing \textsuperscript{\rm \dag},\\
    Zihan Fang\textsuperscript{\rm \P},
    Heming Cui\textsuperscript{\rm \dag},
    Dong Huang\textsuperscript{\rm \#,*}\\
    \textsuperscript{\rm \dag} Department of Computer Science, The University of Hong Kong, Hong Kong, China.\\
    \textsuperscript{\rm \ddag} Department of Electronic Engineering, Shanghai Jiao Tong University, Shanghai, China. \\
    \textsuperscript{\rm \S} Department of Electrical and Electronic Engineering, The University of Hong Kong, Hong Kong, China. \\
    \textsuperscript{\rm \P} Department of Computer Science, City University of Hong Kong, Hong Kong, China. \\
    \textsuperscript{\rm \#} School of Computing, National University of Singapore, Singapore.\\
    \thanks{* Dong Huang is the corresponding author. Email: dhuang@nus.edu.sg}
    \thanks{The work is supported in part by National Key R\&D Program of China (2022ZD0160201), HK RGC RIF (R7030-22), HK RGC GRF (ref No.: 17208223 \& 17204424), a Huawei flagship research grant in 2023, SupernetAI, and the HKU-CAS Joint Laboratory for Intelligent System Software.}
    \thanks{Copyright 2025 IEEE. Published in ICPADS 2025 – the 31st IEEE International Conference on Parallel and Distributed Systems (ICPADS), scheduled for 14-18 December 2025 in Hefei, China. Personal use of this material is permitted. However, permission to reprint/republish this material for advertising or promotional purposes or for creating new collective works for resale or redistribution to servers or lists, or to reuse any copyrighted component of this work in other works, must be obtained from the IEEE. Contact: Manager, Copyrights and Permissions / IEEE Service Center / 445 Hoes Lane / P.O. Box 1331 / Piscataway, NJ 08855-1331, USA. Telephone: + Intl. 908-562-3966.}
}

\maketitle

\begin{abstract}
Deep neural networks (DNNs) are vulnerable to adversarial examples, in which DNNs are misled to false outputs due to inputs containing imperceptible perturbations. Adversarial training, a reliable and effective method of defense, may significantly reduce the vulnerability of neural networks and becomes the de facto standard for robust learning. While many recent works practice the data-centric philosophy, such as how to generate better adversarial examples or use generative models to produce additional training data, we look back to the models themselves and revisit the adversarial robustness from the perspective of deep feature distribution as an insightful complementarity. In this paper, we propose \textit{Branch Orthogonality adveRsarial Training} (BORT) to obtain state-of-the-art performance with solely the original dataset for adversarial training. To practice our design idea of integrating multiple orthogonal solution spaces, we leverage a simple multi-branch neural network and propose a corresponding loss function, branch-orthogonal loss, to make each solution space of the multi-branch model orthogonal. We evaluate our approach on CIFAR-10, CIFAR-100 and SVHN against $\ell_{\infty}$ norm-bounded perturbations of size $\epsilon = 8/255$, respectively. Exhaustive experiments are conducted to show that our method goes beyond all state-of-the-art methods without any tricks. Compared to all methods that do not use additional data for training, our models achieve 67.3\% and 41.5\% robust accuracy on CIFAR-10 and CIFAR-100 (improving upon the state-of-the-art by +7.23\% and +9.07\%). 
\end{abstract}

\begin{IEEEkeywords}
Deep neural network, adversarial training, model robustness, multi-branch network.
\end{IEEEkeywords}

\section{Introduction}
Deep Neural Networks (DNNs) have got great research attention and success in various application scenarios~\cite{lin2024efficient,yuan2025constructing,peng2024sums,lin2024adaptsfl,fang2024automated,lin2025hierarchical,duan2025sample,hu2024accelerating}. However, it is found that DNNs are vulnerable to adversarial examples~\cite{zhang2025robust,duan2025rethinking,du2025dp,zhang2025state}. Leaving the model to produce the wrong output as the attacker wishes poses a significant security risk. Therefore, the vulnerability of DNNs to adversarial attacks is a critical topic in many areas with safety-vital nature, such as autonomous driving~\cite{wang2021interpreting,fang2024ic3m,lin2025hsplitlora}, and medical diagnostics~\cite{de2018clinically,fang2025dynamic,tang2024merit}.

\begin{figure}[t]
\centering
\includegraphics[width=0.8\linewidth]{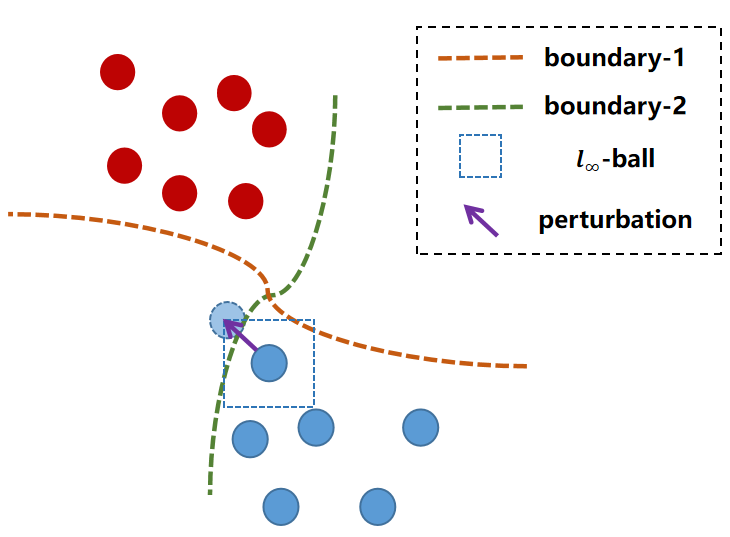}
\caption{The distribution of simplified binary classification models in the solution space, where blue and red dots represents two categories of data samples respectively.}
\label{fig:intro}
\end{figure}

\begin{figure*}
	\begin{minipage}[t]{0.5\linewidth}
		\centering
		\includegraphics[width=0.8\linewidth]{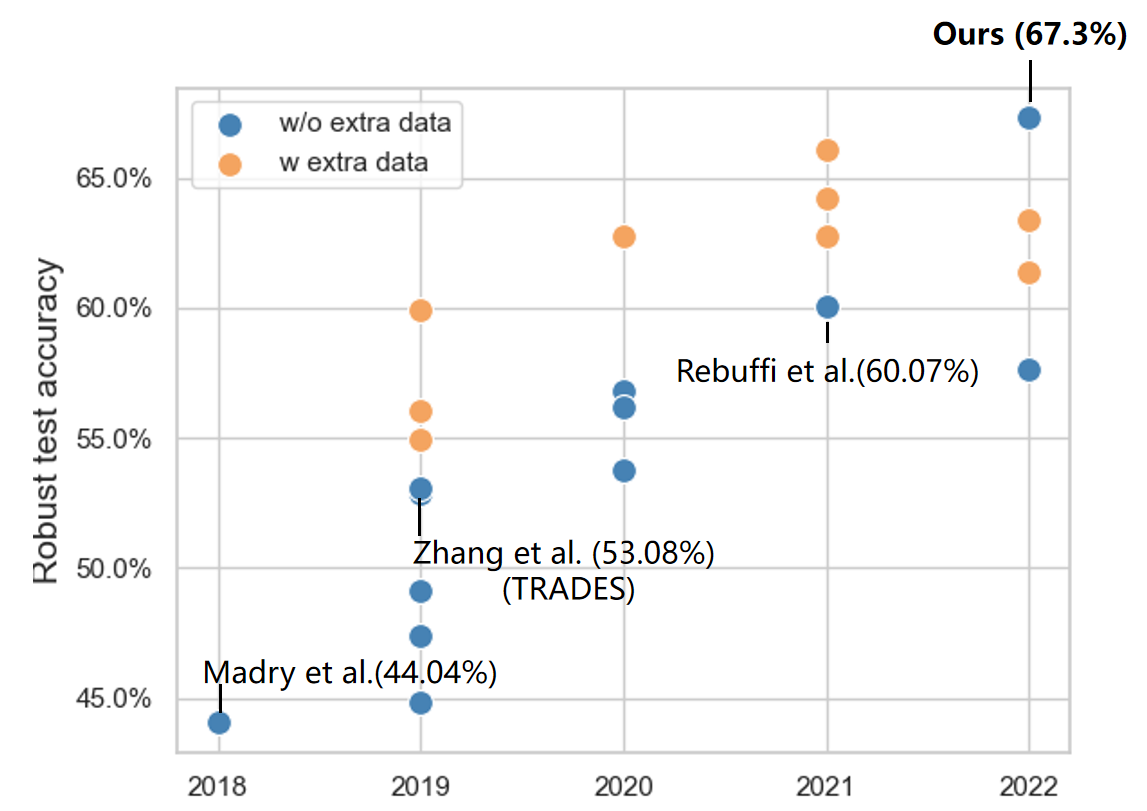}
	\end{minipage}
	\begin{minipage}[t]{0.5\linewidth}
		\centering
		\includegraphics[width=0.8\linewidth]{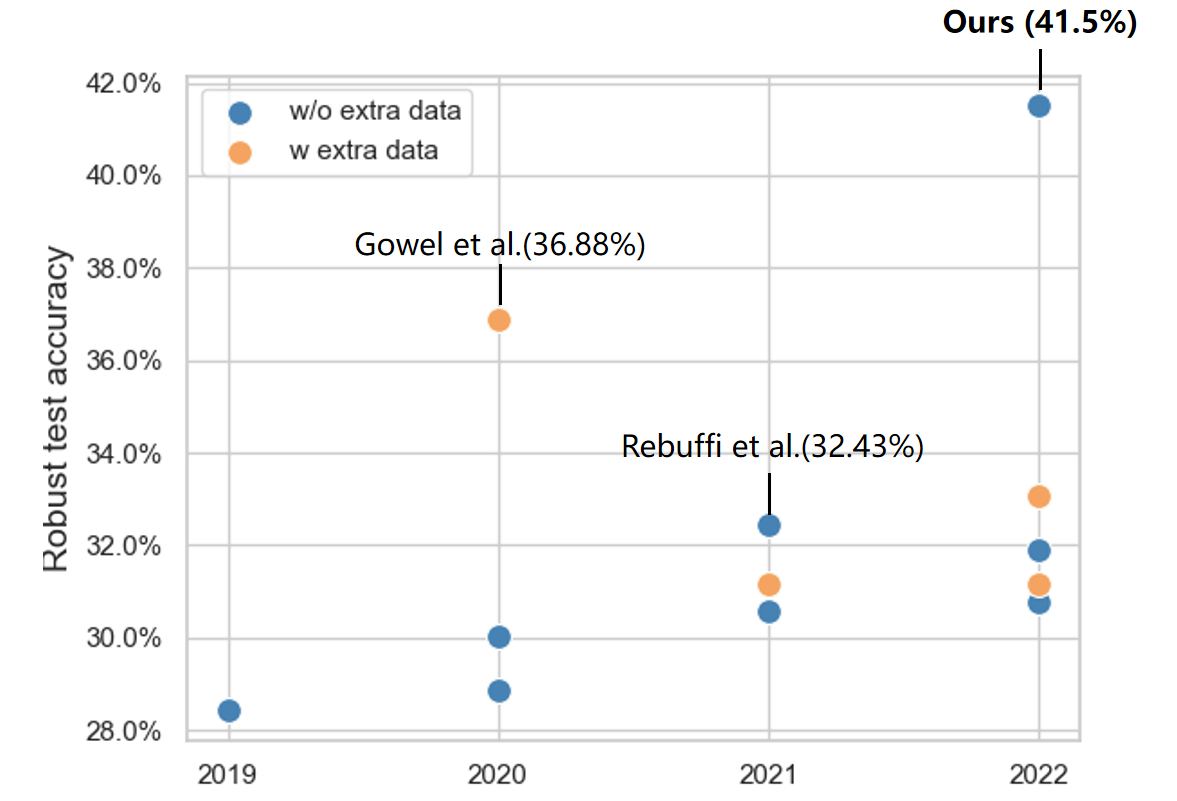}
	\end{minipage}
    \caption{Robust accuracy of models against AutoAttack\cite{croce2020reliable} on CIFAR10 (left) and CIFAR100 (right) dataset.}
    \label{fig:1}
\end{figure*}

Ever since Szegedy et al.~\cite{szegedy2013intriguing} brought this problem to the forefront, many studies have sprung up on generating adversarial examples~\cite{athalye2018synthesizing} and building defenses against these attacks~\cite{jeddi2020learn2perturb}. Among many successful defenses, adversarial training became one of the mainstream and proved to be effective and practical. 
Madry et al.~\cite{madry2017towards} first introduced projected gradient descent (PGD) to generate adversarial examples for training, achieving model robustness and invariance to input perturbations. As a standard for training adversarially robust models, a variety of modifications over their implementation have been proposed, such as Lipschitz regularization~\cite{yan2018deep}, computationally efficient adversarial training~\cite{wong2020fast}, utilization of unlabeled data~\cite{carmon2019unlabeled} and better trade-offs between robustness and natural generalization~\cite{zhang2020attacks}.

However, in recent years, the improvement in robustness of the model seems to be less effective and practical. To further advance the model robustness, Carmon et al.~\cite{carmon2019unlabeled}, Najafi et al.~\cite{najafi2019robustness} and Zhai et al.~\cite{zhai2019adversarially} pioneered the use of additional data (under both supervised and semi-supervised settings). Gowal et al.~\cite{gowal2020uncovering} trained a model on CIFAR-10 against $\ell_{\infty}$ perturbations of size $\epsilon = 8/255$ with a maximum enhancement of $+8.73\%$ robust accuracy when using additional unlabeled data, uncovering the importance of additional datasets for robustness of the model obtained from adversarial training.

Although the methods listed above have proven to be very effective and continually advance the boundaries of model robustness, these works mainly focus on the data for adversarial training but the model. As we observed, they are all subject to the same restrictions, namely, that the solution space of the model leveraged is onefold and immutable, leaving it open to white-box attacks. It is also evident from the experimental observations that as one increases the scope limitation of perturbation, they produce a significant decrease in robust accuracy.

We revisit the issue of model robustness under attack from the perspective of solution (i.e. representation) space. As demonstrated in Fig. {\ref{fig:intro}}, a single classification boundary can be easily and successfully attacked by a perturbation bounded in the $\ell_\infty$-ball, yet another classification boundary with a different distribution of solution space can still correctly classify the perturbed example. Therefore, we come up with the idea of using a neural network with several branches, each of which is a standard classification model(e.g. ResNet, WRN), to perform this integration of classification boundaries.
To be more rigorous, the multi-branch structure we leverage is just a simple design paradigm for implementing multiple solution spaces, where the \textit{branch} itself can stand for a broader concept, such as solution space.
And it intuitively stands to reason that the more distinct these classification boundaries are, the more complementary they are, and the more effective they will work together as a whole. However, realizing solution space orthogonality is quite challenging because the process of optimizing network parameters using stochastic gradient descent is very uncertain. 
To make the different classification boundaries corresponding to the different branches vary widely, a novel loss function for supervision is necessary. Therefore, we propose branch-orthogonal loss, derived from the idea of using cosine similarity to evaluate the orthogonality between vectors, to let the newly trained branches optimize in a different direction from the already trained ones as much as possible.

We propose \textit{Branch Orthogonality adveRsarial Training} (BORT), a method to build robust models with the multi-branch structure that holds multiple orthogonal solution spaces, performing as a complementarity to the current prevailing data-centric design philosophy. Our method surpasses all current methods with or without additional datasets in robust accuracy, as listed in Fig. {\ref{fig:1}}. To conclude, our contributions can be folded as follows:

\begin{itemize}
    \item {We revisit the model robustness from the perspective of solution space and reveal the effectiveness of multiple orthogonal solution spaces in improving the overall model robustness.}
    \item {A multi-branch model structure and its training pipeline are introduced to build models with state-of-the-art robustness against adversarial attacks. (Robust accuracies of 67.3\% on CIFAR-10 and 41.5\% on CIFAR-100 against $\ell_{\infty}$ perturbations AutoAttack, with an improvement of about +7.23\% and +9.07\% compared to the current state-of-the-art methods without extra training data.)}
    \item {We propose the branch-orthogonal loss, which provides supervision between multiple branches to achieve multi-branch solution space orthogonality.}
    \item {Despite of the need of training multiple branches, our method remains a moderate training cost since no extra data is used for better performance (i.e., neither generative models nor extra sizable datasets).}
\end{itemize}
\section{Related Work}

\begin{figure*}[htbp]
\centering
\includegraphics[width=0.75\linewidth]{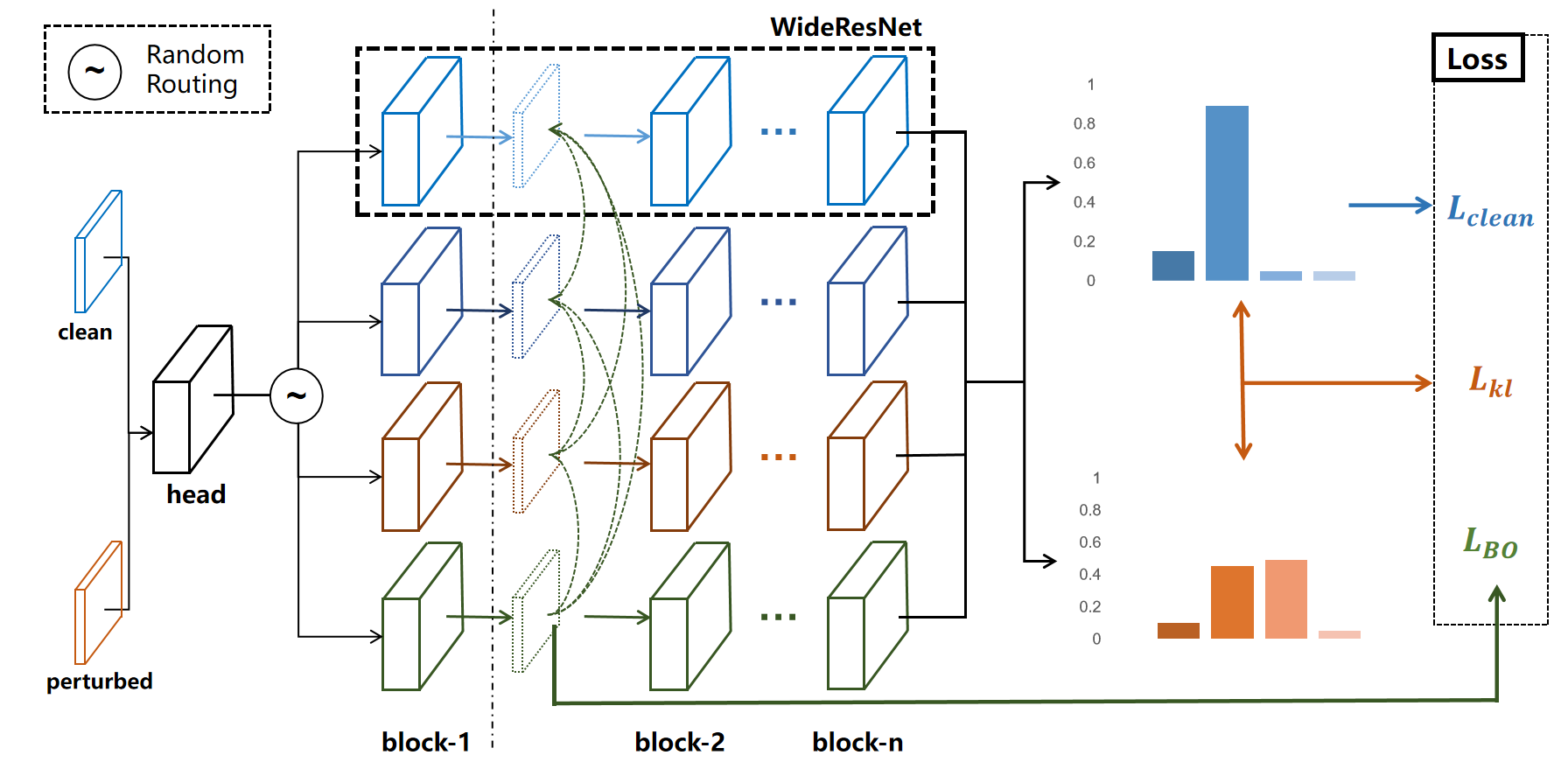}
\caption{Overview}
\label{fig:overview}
\end{figure*}

\subsection{Adversarial Attacks}

Research on DNN robustness primarily involves adversarial attacks and defenses. Adversarial attacks generate subtle perturbations to input data, causing models to misclassify. These attacks are categorized as white-box (with full model access) or black-box (without model access), and as targeted (forcing a specific misclassification) or untargeted (any misclassification). Goodfellow, Shlens, and Szegedy~\cite{goodfellow2014explaining} proposed the Fast Gradient Sign Method (FGSM) which generates adversarial examples. It was followed by R+FGSM~\cite{tramer2017ensemble}, which adds a randomization step, and the Basic Iterative Method (BIM)~\cite{kurakin2018adversarial}, which takes multiple smaller gradient steps. In this paper, we mainly focus on AutoAttack~\cite{croce2020reliable}, a widely used robustness benchmark that combines four complementary attacks, to evaluate and compare our method.

\subsection{Adversarial Training as a defense}

In response to the rapid development of adversarial attacks, researchers have proposed various defense methods, mainly categorized as certified and empirical defenses. Certified defenses use optimized objectives to guarantee robustness~\cite{hein2017formal,raghunathan2018certified,sinha2017certifiable}, while empirical defenses train models with adversarial samples.

Adversarial training is one of the notable defense methods as an empirical defense, which is also adopted in this paper to build robust models. And its basic idea can be expressed as a min-max optimization problem:
\begin{equation}
    \theta^* = \mathop{\arg\min}\limits_{\theta} \mathbb{E}_{(x,y)\in X}\left[\mathop{\max}\limits_{\delta \in [-\epsilon, \epsilon]}\ell(f_{\theta}(x+\delta), y)\right],
\end{equation}
where the $f_\theta$ is a DNN parameterized by $\theta$, $X$ stands for the training dataset and $\ell$ represents the loss function. And perturbations $\delta$ are bounded into the $\epsilon$-ball. 

PGD~\cite{madry2017towards} is a widely used method for generating adversarial perturbations by iteratively updating inputs and projecting them onto an $l_p$-norm ball, making it more effective than single-step attacks like FGSM~\cite{goodfellow2014explaining}.

Beyond adversarial training, various non-certified defenses (e.g., distillation~\cite{papernot2016distillation}, activation pruning~\cite{dhillon2018stochastic}, input transformation~\cite{guo2017countering}) have been proposed, but many suffer from \textit{obfuscated gradients}\cite{athalye2018obfuscated} and can be bypassed by advanced attacks. In contrast, adversarial training methods such as PAT\cite{madry2017towards}, TRADES~\cite{zhang2019theoretically}, and MART~\cite{wang2019improving} remain effective against both white-box and black-box attacks. Recent work by Gowal et al.\cite{gowal2020uncovering} revealed a robustness gap between models trained with and without extra data like 80M-TI. Follow-up studies\cite{gowal2021improving, rebuffi2021fixing} showed that even synthetic data from generative models and various augmentation strategies can further enhance robustness.

\section{Proposed Method}

We propose a novel framework, BORT, by introducing a multi-branch neural network and a new loss term for orthogonality learning. We first introduce the pipeline of our framework in Sec.~\ref{III.A} and then present our novel branch-orthogonal loss in Sec.~\ref{III.B} and our optimization formulation in Sec.~\ref{III.C} followed by the training pipeline in Sec.~\ref{III.D}.

\textbf{Notation.}
$(x,y)\in \mathcal{X}$ is the natural dataset, and $x'$ signifies perturbed examples. $f^{k}$ denotes any of the $K$ branches and $f_i^{k}$ especially stands for the i-th `block' of branch $k$. $\delta$ indicates perturbations bounded in the $\epsilon$-ball, where $\epsilon$ represents the numerical limit of the perturbation.

\begin{algorithm}[t]
	\renewcommand{\algorithmicrequire}{\textbf{Input:}}
	\renewcommand{\algorithmicensure}{\textbf{Output:}}
	\caption{Training Pipeline}
	\label{alg1}
	\begin{algorithmic}[1]
    \REQUIRE Model $f$,\ data $x$,\ labels $y$,\ num\ of\ branches $K$,\ number\ of\ epochs $epochs$,\ PGD algorithm $\mathrm{PGD}()$
    
     \FOR{$k=0$ \TO $k=K$} 
     \STATE $\lambda_1 , \lambda_2 \leftarrow 3$
     \IF {$k > 0$}
     \STATE  $f.head.parameters().requires\_grad \leftarrow False$
     \ENDIF
     
     \FOR{$e=0$ \TO $e=epochs$}
     \IF{$e > 40$}
     \STATE $\lambda_1 , \lambda_2 \leftarrow 1$
     \ENDIF
     \STATE $\hat{y} \ \, \leftarrow f^{k}(x) $
     \STATE {$x' \leftarrow \mathrm{PGD}(f^i, x, y)$}
     \STATE $\hat{y}' \leftarrow f^{k}(x') $

     \FOR{$j=0$ \TO $j=k-1$}
     \STATE $\mathcal{L}_{\mathrm{BO}} += \frac{1}{k}\  | \mathrm{cos\_sim}(f_1^{j}(x'), f_1^k(x')) |$
	 \ENDFOR
	 
	 \STATE $\mathcal{L}_{\mathrm{clean}} \leftarrow \mathrm{CE}(\hat{y},y)$
     \STATE $\mathcal{L}_{\mathrm{KL}} \leftarrow \mathrm{KL}(\hat{y}',\hat{y})$
	 
	 \STATE $loss \leftarrow \mathcal{L}_{\mathrm{clean}}+ \lambda_1 \mathcal{L}_{\mathrm{KL}}+\lambda_2 \mathcal{L}_{\mathrm{BO}} $ 
	 
	 \STATE $loss.backward()$
	 \STATE $f^k.update()$
	 
	\ENDFOR
	\ENDFOR
	 
	\end{algorithmic}  
	\label{alg:training}
\end{algorithm}
\subsection{Multi-branch Neural Network}
\label{III.A}
The overview pipeline of our proposed method can be roughly concluded in Fig. {\ref{fig:overview}}. The clean and perturbed examples will first go 
through a shared head, consisting of a 2D convolutional layer. Then, the generated features will be randomly routed to one of four branches, each of which is an independent WideResNet model. We adopt the commonly used three settings with different model scales: WRN-28-10, WRN-34-10, and WRN-70-16, respectively. The possibility of routing to each branch is equal. Our multi-branch structure shares a greater model capacity and representation ability, while the inference time and computational cost stay consistent with an individual WRN model. The overall performance of our method compared to state-of-the-arts will then be exhaustively discussed in Sec.~\ref{IV}.

\textbf{Randomness might be enough.}
Routing to one of $K$ branches randomly is a thoroughly non-differentiable operation, so that white-box attacks that are based on optimization will be completely invalidated. To be more precise, every iteration for generating perturbations can only target a specific branch of the network, not the entire. Predictably, if we use some differentiable approach for branch selection, the multi-branch network will still be very vulnerable, and the multi-branch design will then become redundancy.

\subsection{Branch-orthogonal Loss}
\label{III.B}
To guide the learning of multi-branch models, we heuristically propose an additional loss term to make the solution spaces of different branches orthogonal, which we believe will maximize the model robustness. When training the last branch ($k$-th in $K$ branches) of our model, its mathematical form is as follows:
\begin{align}
\mathcal{L}_{\mathrm{BO}} &= \frac{1}{k-1}\sum_{j=1}^{k-1} | \mathrm{cos\_sim}(f_1^{j}(x'), f_1^k(x')) \lvert \notag \\
 &= \frac{1}{k-1}\sum_{j=1}^{k-1} | \frac{f_{1}^{k}(x')  f_1^{j}(x')}{\max(\Vert f_1^{k}(x') \Vert \cdot \Vert f_1^{j}(x') \Vert, \epsilon)} \rvert,
\end{align}
where $\mathrm{cos\_sim}$ denotes the cosine similarity, and $\epsilon$ is a sufficiently small and positive number.

\subsection{Optimization Formulation}
\label{III.C}

\textbf{Loss for inner maximization.}
Retrospective of the mini-max optimization problem in adversarial training, the inner maximization is for generating adversarial examples $x'$ from clean examples $x$. Here, we use the commonly used the cross-entropy loss to find adversarial examples as follows:
\begin{equation}
    x' = x + \mathop{\arg\max}\limits_{\delta\in[-\epsilon,\epsilon]}{\mathrm{CE}(f^{k}(x+\delta), y)}.
\end{equation}
where $\mathrm{CE}$ denotes the cross-entropy loss.

\textbf{Loss for outer minimization.}
The outer minimization problem is to allow the model to correctly classify adversarial examples.
To obtain a better trade-off between the model robustness and its accuracy on clean data, we use the loss design proposed in TRADES\cite{zhang2019theoretically}. In addition to the branch-orthogonal loss we introduced for orthogonality learning, the loss is composed of two parts: the first encourages the natural error (on clean data) to be minimized by applying the cross-entropy loss between the prediction of clean example $f^{k}(x)$ and labels $y$, and the second acts like a regularization term optimizing the difference between the prediction
of clean example $f(x)$ and that of adversarial example $f^{k}(x')$. And there is a hyper-parameter balancing the importance of the two parts.

\textbf{End-to-end optimization.}
The full version of loss that we use to optimize networks is presented here, combining all of the aforementioned loss terms. It consists of three distinct components, each of which serves a different purpose: the cross-entropy loss to fit our model to clean examples, the Kullback-Leibler loss to close the gap between prediction results for clean and perturbed examples, and our newly proposed Branch-orthogonal loss to make the solution space of each branch orthogonal.
\begin{align}
&\mathcal{L}_{\mathrm{clean}} = \mathrm{CE}(f^{k}(x),y), \\
&\mathcal{L}_{\mathrm{KL}} = \mathrm{KL}(f^{k}(x'),f^{k}(x)), \\
&\mathcal{L}_{\mathrm{BO}} = \frac{1}{k-1}\sum_{j=1}^{k-1} | \mathrm{cos\_sim}(f_1^{j}(x'),f_1^k(x')) | ,\\
&\mathcal{L}_{\mathrm{BORT}} = \mathcal{L}_{\mathrm{clean}}+ \lambda_1 \mathcal{L}_{\mathrm{KL}}+\lambda_2 \mathcal{L}_{\mathrm{BO}},
\end{align}
where $\mathrm{KL}$ denotes the Kullback-Leibler loss, $\lambda_1$ and $\lambda_2$ are tuning parameters balancing the importance among the three loss function.

\subsection{Training Pipeline}
\label{III.D}

The overall training process of our proposed method is illustrated in Algorithm{~\ref{alg:training}}, where the $K$ branches of the network are trained separately in a serial fashion. 
We use the PGD algorithm to generate adversarial examples under the setting of $\epsilon=8/255, \alpha=0.01$ and $\mathrm{steps}=10$. The branch-orthogonal loss is calculated between the currently trained branch with all branches that have been trained previously then averaged with $k$. For example, when training branch No.2 (indexing from 0), branch-orthogonal loss will be calculated in branch No.0 with branch No.2, and No.1 with No.2.  

\section{Experiment}
\label{IV}
\subsection{Evaluation Setting}

\begin{table*}[h!]
\small
\centering
\setlength{\tabcolsep}{4.5mm}{
\begin{tabular}{l l l c c}
\toprule
\textbf{Dataset} & \textbf{Defense Method} & \textbf{Model} & \textbf{Clean} & \textbf{AutoAttack} \\
\midrule

\multirow{13}{*}{\textbf{CIFAR-10}}
& Wu, Xia, and Wang\cite{wu2020adversarial} & WRN-28-10(RST-AWP) & 88.3\% & 60.0\% \\
& Gowal et al.\cite{gowal2020uncovering} & WRN-70-16 & 85.3\% & 57.2\% \\
& Rebuffi et al.\cite{rebuffi2021data} & WRN-28-10 & 86.1\% & 57.5\% \\
& Rebuffi et al.\cite{rebuffi2021data} & WRN-34-10 & 86.2\% & 58.1\% \\
& Rebuffi et al.\cite{rebuffi2021data} & WRN-70-16 & 87.3\% & 60.1\% \\
& Sehwag et al.\cite{sehwag2021robust} + 10M DDPM & WRN-34-10 & 86.7\% & 60.3\% \\
& Pang et al.\cite{pang2022robustness} + 1M DDPM & WRN-28-10 & 88.6\% & 61.0\% \\
& Pang et al.\cite{pang2022robustness} + 1M DDPM & WRN-70-16 & 89.0\% & 63.4\% \\
& Jia et al.\cite{jia2022adversarial} & WRN-34-10(LAS-AWP) & 87.7\% & 55.5\% \\
& Jia et al.\cite{jia2022adversarial} & WRN-70-16 & 85.7\% & 57.6\% \\
& BORT(Our) & WRN-28-10 & 87.5\% & \textbf{65.8\%} \\
& BORT(Our) & WRN-34-10 & 87.7\% & \textbf{66.3\%} \\
& BORT(Our) & WRN-70-16 & 88.3\% & \textbf{67.3\%} \\
\midrule

\multirow{10}{*}{\textbf{CIFAR-100}}
& Wu, Xia, and Wang\cite{wu2020adversarial} & WRN-34-10 & 60.4\% & 28.9\% \\
& Gowal et al.\cite{gowal2020uncovering} & WRN-70-16 & 60.9\% & 30.0\% \\
& Rebuffi et al.\cite{rebuffi2021data} & WRN-28-10 & 63.0\% & 29.8\% \\
& Rebuffi et al.\cite{rebuffi2021data} & WRN-70-16 & 65.8\% & 32.4\% \\
& Sehwag et al.\cite{sehwag2021robust} + 1M DDPM & WRN-34-10 & 65.9\% & 31.2\% \\
& Pang et al.\cite{pang2022robustness} + 1M DDPM & WRN-28-10 & 63.7\% & 31.1\% \\
& Pang et al.\cite{pang2022robustness} + 1M DDPM & WRN-70-16 & 65.6\% & 33.1\% \\
& Jia et al.\cite{jia2022adversarial} & WRN-34-10 & 64.9\% & 30.8\% \\
& BORT(Our) & WRN-28-10 & 63.7\% & \textbf{38.2\%} \\
& BORT(Our) & WRN-34-10 & 64.2\% & \textbf{39.0\%} \\
& BORT(Our) & WRN-70-16 & 63.7\% & \textbf{41.5\%} \\
\midrule

\multirow{6}{*}{\textbf{SVHN}}
& Wu, Xia, and Wang\cite{wu2020adversarial} & WRN-34-10 & 92.8\% & 58.7\% \\
& Gowal et al.\cite{gowal2020uncovering} & WRN-28-10 & 92.9\% & 57.3\% \\
& Gowal et al.\cite{gowal2020uncovering} & WRN-34-10 & 93.1\% & 58.0\% \\
& Rebuffi et al.\cite{rebuffi2021data} & WRN-28-10 & 94.5\% & 57.3\% \\
& BORT(Our) & WRN-28-10 & 93.1\% & \textbf{59.7\%} \\
\bottomrule
\end{tabular}}
\caption{Comparison of state-of-the-art adversarial defense methods with clean and perturbed-data accuracy on CIFAR-10, CIFAR-100, and SVHN dataset under AutoAttack with $l_{\infty}$ perturbations of size 8/255.}
\vspace{-0.2cm}
\label{tab:cifar10}
\end{table*}

\textbf{Datasets and adversarial attacks.}
Three publicly available benchmark datasets are used. The CIFAR-10 dataset consists of 60,000 32x32 color images of 10 classes, with 6,000 images per class. There are 50,000 training images and 10,000 test images. The CIFAR-100 dataset is like CIFAR-10, except that it has 100 classes, each containing 600 images, of which 500 are training images and 100 are test images. The Street View House Number (SVHN) Dataset is derived from Google street view house numbers, each image contains a set of '0-9' Arabic numbers.
The training set contains 73257 digits and the test set contains 26032 digits.

FGSM, PGD, and AutoAttack are utilized to evaluate our proposed method along with state-of-the-arts. Testing robustness against PGD is done in 40/100 iterations with the maximum $\epsilon=8/255$ and step sizes of $0.01$ for each iteration. FGSM attack also uses the same 8/255 bound for perturbation. We leverage the implementation of AutoAttack from torchattacks repository~\cite{kim2020torchattacks}.

\textbf{Model training details.}
We use WideResNets as our base models: WRN-28-10, WRN-34-10 and WRN-70-16~\cite{zagoruyko2016wide}, respectively. Each branch is trained for 50 epochs using stochastic gradient descent (SGD) as the optimizer, with an initial learning rate of 0.1 that decays to 0.01 at the 40th epoch. The hyper-parameters are configured as $\lambda_1 = \lambda_2 = 3$ during the first 40 epochs, followed by $\lambda_1 = \lambda_2 = 1$ for the remaining 10 epochs.

\begin{table*}[t]
\centering
\small
\setlength{\tabcolsep}{6mm}{
\begin{tabular}{lccccc}
\toprule
\textbf{Model}  & \textbf{Clean} & \textbf{FGSM} & \textbf{PGD40} &\textbf{PGD100} &\textbf{AutoAttack}\\
\midrule
BORT-WRN-28-10-1&	87.1\%&	52.0\%&	52.0\%&	51.9\%&	51.5\%    \\
BORT-WRN-28-10-2&	86.1\%&	55.3\%&	56.7\%&	55.9\%&	59.5\% \\
BORT-WRN-28-10-4&	87.5\%&	64.7\%&	59.4\%&	59.3\%&	65.8\% \\
\midrule
\midrule
BORT-WRN-70-16-1&	87.3\%&	61.1\%&	55.8\%&	55.7\%&	54.1\% \\
BORT-WRN-70-16-2&	87.6\%&	66.2\%&	58.3\%&	57.9\%&	62.2\% \\
BORT-WRN-70-16-4&	88.3\%&	67.7\%&	60.5\%&	60.4\%&	67.3\% \\
\bottomrule
\end{tabular}}
\caption{The performance of single-branch, two-branch and the complete four-branch model on the CIFAR10 dataset facing adversarial attacks FGSM, PGD and AutoAttack, respectively.}
\label{tab:numofbranches}
\end{table*}

\begin{table*}[t]
\centering
\small
\setlength{\tabcolsep}{3mm}{
\renewcommand{\arraystretch}{1.3}
\begin{tabular}{c| c c c | c c c c c}
\hline
\multirow{2}{*}{\textbf{Methods}} & 
\multicolumn{3}{c|}{\textbf{Loss}} & \multicolumn{5}{c}{\textbf{Attacks}}\\
\cline{2-9}

 &$\mathcal{L}_{\mathrm{clean}}$ &$\mathcal{L}_{\mathrm{KL}}$ &$\mathcal{L}_{\mathrm{BO}}$& Clean & FGSM & PGD40 & PGD100 & AutoAttack \\
\hline

Vanilla	& \checkmark& & & 93.8 & 8.2 & 0 & 0 & 0 \\
PAT	& & \checkmark(CE)& &86.5	&60.5	&52.5	&52.5	&50.5\\
TRADES	& \checkmark&\checkmark & &86.6	&58	&51.1	&51	&49.8\\
TRADES-4	& \checkmark& \checkmark& &86.3	&64.7	&58.7	&58.7	&64.9\\
BORT(WRN-28-10)	&\checkmark &\checkmark &\checkmark &87.5	&64.7	&59.4	&59.3	&65.8\\
BORT(WRN-70-16)	&\checkmark &\checkmark &\checkmark &88.3	&67.7	&60.5	&60.4	&67.3\\

\bottomrule

\end{tabular}}
\caption{Ablation studies of loss terms. Vanilla: Models trained entirely with clean samples. TRADES-4: The same model structure as our method proposed, but the branches are trained according to TRADES\cite{zhang2019theoretically}(differ in loss function). }
\label{tab:compare}
\end{table*}


\subsection{Overall Performance}
To illustrate the effectiveness of the proposed BORT, we train various models using this framework and evaluate their robustness against different adversarial attack algorithms. Furthermore, the proposed method is compared with different state-of-the-art approaches~\cite{wu2020adversarial,gowal2020uncovering,jia2022adversarial,sehwag2021robust,pang2022robustness,rebuffi2021data}. The detailed experiment results are illustrated in Table \ref{tab:cifar10}. For a fair comparison, only the results that do not use the additional dataset~(80M-Ti for example) are presented in the table, although we still outperform the results using it. The 80M-Ti dataset has been taken down by the author team, given that it may contain offensive and biased content. Likewise, we do not try to train stronger models utilizing this dataset as additional training data.

As shown in Table~\ref{tab:cifar10}, our method achieves maximum robustness improvements of \textbf{+7.23\%} on CIFAR-10, \textbf{+9.07\%} on CIFAR-100, and \textbf{+1.00\%} (WRN-28-10 only) on SVHN over the best results among methods that do not use additional datasets. Our method also surpasses those trained with extra datasets: improvement of \textbf{+3.95\%} and \textbf{+8.45\%} in CIFAR-10 and CIFAR-100, respectively. In a longitudinal comparison of our method, it is consistent with the intuition that using a larger base model leads to an overall performance improvement. This improvement, however, is more minor than that provided by our proposed method.

With the same base model, we achieve the best results on robustness compared to all previous methods. It’s worth mentioning that our method outperforms all current SOTA methods with base model WRN-70-16 by a significant gap of \textbf{+7.23\%}(extra data SOTA \textbf{+3.95\%}) in CIFAR-10 and \textbf{+9.07\%}(extra data SOTA \textbf{+8.45\%}) in CIFAR-100. Even with our lightest base model WRN-28-10, we still have a huge performance improvement of \textbf{+5.76\%}(extra data SOTA \textbf{+4.76\%}) in CIFAR-10 and \textbf{+8.4\%}(extra data SOTA \textbf{+7.12\%}) in CIFAR-100. 

\subsection{Ablation Studies}


\begin{figure}[t]
    \centering
    \begin{subfigure}[b]{0.95\linewidth}
        \centering
        \includegraphics[width=\linewidth]{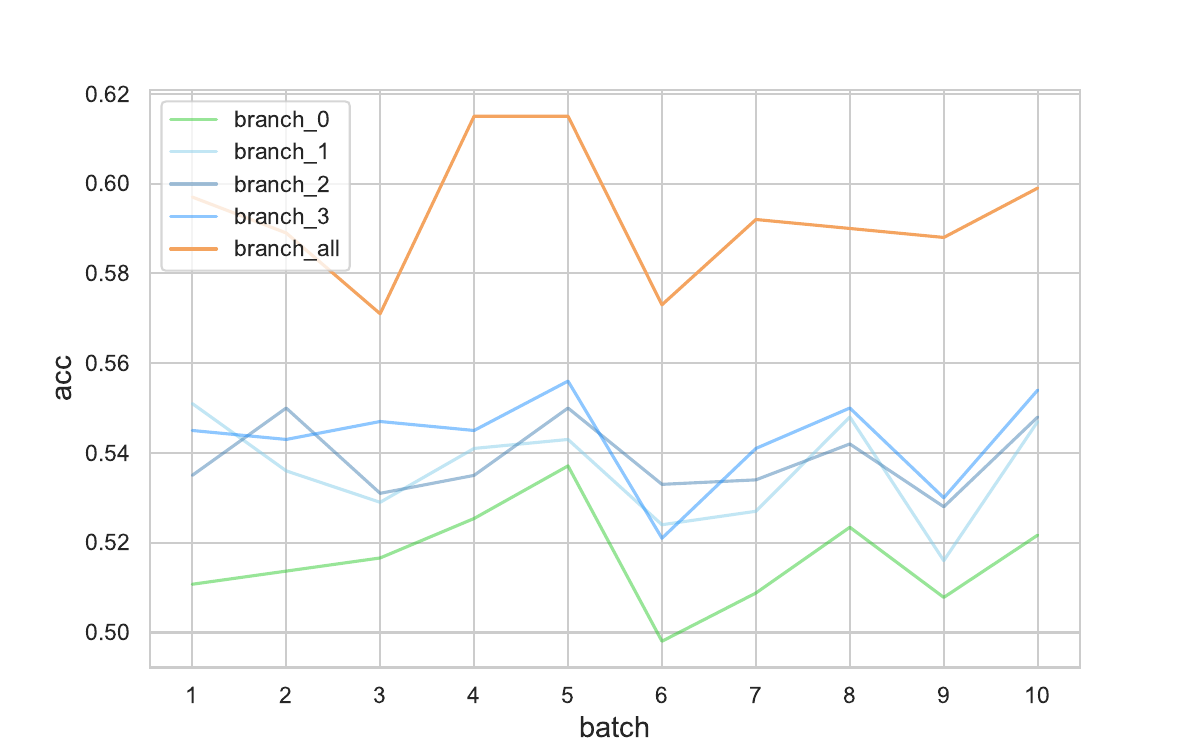}
        \caption{BORT}
        \label{fig:a}
    \end{subfigure}
    \begin{subfigure}[b]{0.95\linewidth}
        \centering
        \includegraphics[width=\linewidth]{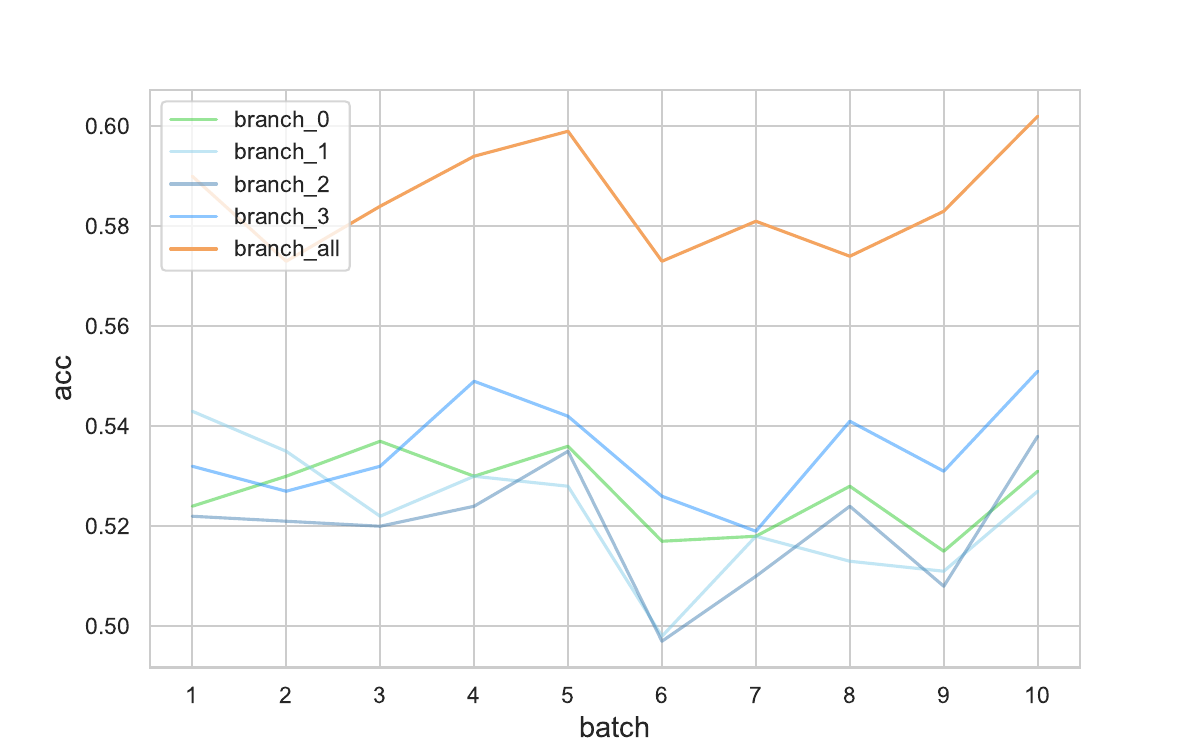}
        \caption{TRADES-4}
        \label{fig:b}
    \end{subfigure}
    \caption{Evaluation results for individual branches on evenly segmented test data batches of CIFAR-10. The model used here is WRN-28-10. Each individual branch is attacked with PGD25 and the complete network with PGD100.}
    \label{fig:every-branch}
\end{figure}
We study the effect of altering the number of branches on model overall robustness, results are listed in Table \ref{tab:numofbranches}. We evaluate the performance of single-branch, two-branch, and the complete four-branch model on the CIFAR-10 dataset facing adversarial attacks FGSM, PGD, and AutoAttack, respectively. The robust accuracy improvement brought by adding more branches is obvious, which is independent of the base model used. It is intuitively consistent that we can achieve higher model robustness by combining multiple branches that hold orthogonally distributed solution spaces. This fact is also evident in the data listed in Fig. {\ref{fig:every-branch}}. 

We are also inspired by the results listed in Fig. \ref{fig:every-branch} to revisit BORT from a different perspective. Expectedly, PGD-100 will iterate 25 times on each branch separately. However, the robust accuracy, with PGD-100 to generate perturbations, obtained from evaluating the complete network is significantly higher than that obtained from evaluating individual branches with PGD-25. We assume that this is due to the random routing mechanism, where PGD does not iterate over a single branch, but takes the results of iterations on other branches as the starting point for optimization, resulting in sub-optimal perturbations that are generated for performing an attack. The iterative optimization directions on different branches may also incur confrontations.

With our proposed branch-orthogonal loss, the complementarity of all branches is prone to be fully exploited by providing supervision to have the solution space of each branch orthogonal. Contrarily, in TRADES-4, the solution space distribution of each branch might be different, but is completely arbitrary and not guaranteed. By fixing the branches chosen at each inference, we evaluate the accuracy of each branch of the model trained by the BORT and TRADES, respectively. Fig. \ref{fig:every-branch}(b) shows that, if there is no Branch-orthogonal loss, each branch has nearly equivalent performance and appears to function similarly. In contrast, with Branch-orthogonal loss that plays a role of guidance, branch No.1/2/3 outperforms branch No.0 by about +2\%, as illustrated in Fig. \ref{fig:every-branch}(a). Overall speaking, the robust accuracy of BORT relative to TRADES-4 is higher by about 0.9\% under AutoAttack and 0.6\% under PGD100 (Table \ref{tab:compare}), with WRN-28-10 as the base model. For a once-iteration attack like FGSM, BO loss is of little help, resulting in consistent performance of TRADE-4 and BORT.

\begin{figure}[h!]
\centering
\includegraphics[scale=0.3]{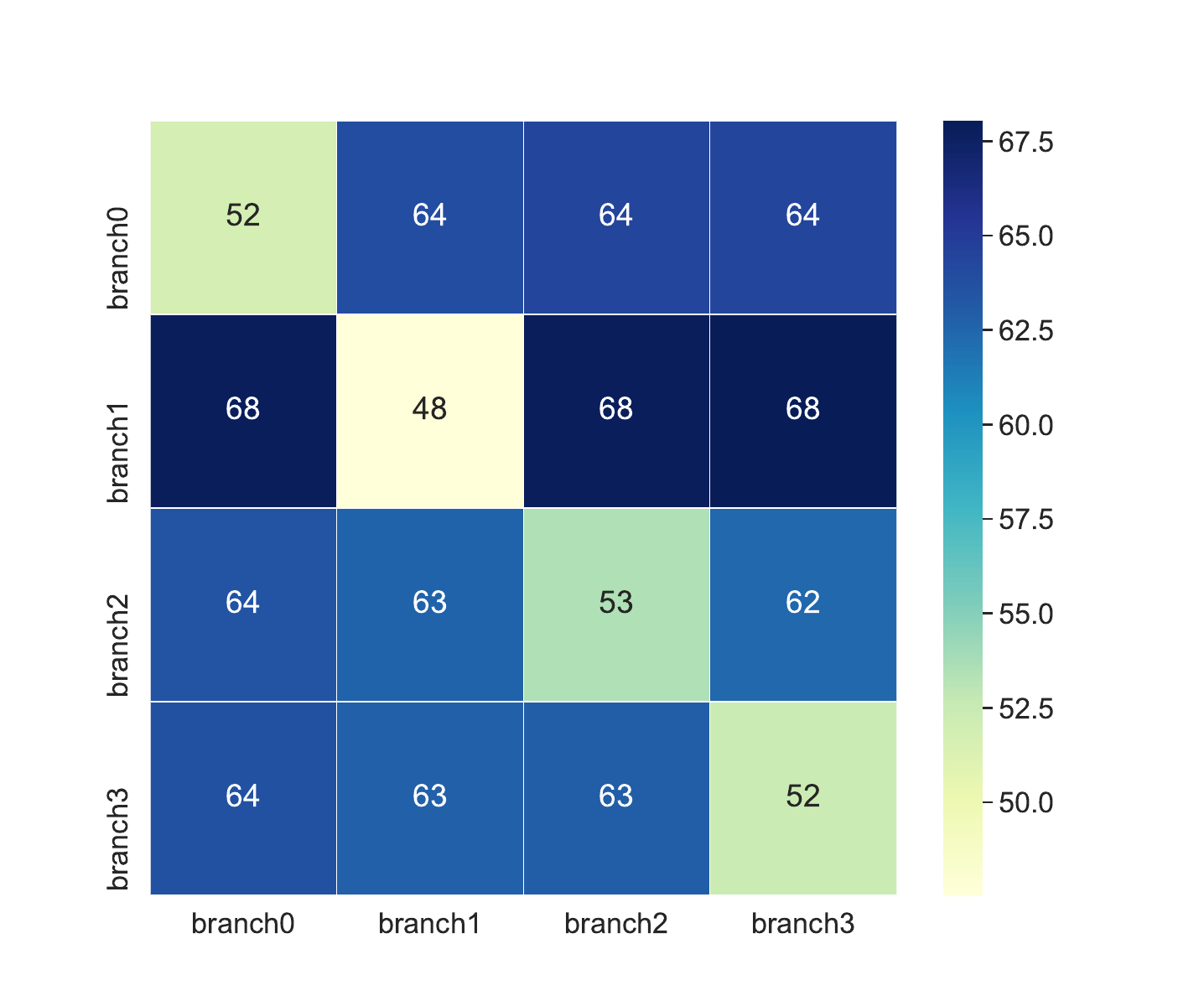}
\caption{Results of transfer attacks among the four branches. The numbers and their corresponding color representations robust accuracy against PGD-100 attack. The dataset and base model we use are CIFAR-10 and BORT-WRN-28-10 respectively.}
\label{fig:transfer}
\end{figure}

We also design an experiment to visualize the difference in the solution space of each branch of BORT and to show the effectiveness of this difference in confronting adversarial attacks. We perform transfer attacks in four branches, i.e., using perturbations generated by each branch to attack other branches. Visualized experiment results are shown in the form of heatmap as in Fig. \ref{fig:transfer}. The lighter the color indicates that the attack is more likely to successfully interfere with the model classification results, and vice versa. A transfer attack yields a significantly lower attack success rate (i.e. higher robust accuracy), compared to each branch attacking itself. This is due to the orthogonality of the solution space of each branch, which prevents the perturbation from having an efficacious effect on the other branches, which is consistent with our motivation illustrated in Fig. \ref{fig:intro}. Another intriguing fact demonstrated in the results is that a less robust model(branch-1) may tend to generate weaker perturbations when performing transfer attacks.

To further validate the effectiveness of the proposed Branch-orthogonal loss, a loss term to encourage the solution space of each branch orthogonal, we retrain a model completely independent of the BORT model, through which we implement the transfer attacks on four branches separately. The experiment results along with more detailed numbers in Table \ref{tab:bo_effect}. From the table, it is clear that perturbations from TRADES-0 have a comparable impact on branches 0 and 3, suggesting that the solution spaces of TRADES-0 and TRADES-3  without Branch-orthogonal loss for training are fairly similar. This similarity in the solution space will degrade the apparent boost brought by integrating multiple solution spaces. The finding further demonstrates that Branch-orthogonal loss can minimize transfer attacks across branches of BORT by making the solution space of each branch as orthogonal as feasible, whereas the solution space produced through TRADES training without Branch-orthogonal loss is not at all guaranteed.

\begin{table}[t]
\small
\centering
\begin{tabular}{c|cccc}
\hline
 & branch-0&  branch-1&  branch-2&  branch-3\\
\hline
TRADES-0& 51.95\%& 61.58\%& 62.73\%& 52.84\%\\
BORT-0 & 51.81\%& 63.92\%& 64.39\%& 64.39\%\\
BORT-1 & 67.85\%& 47.54\%& 67.86\%& 68.04\%\\
BORT-2 & 63.50\%& 62.62\%& 53.46\%& 62.39\%\\
BORT-3 & 63.66\%& 62.83\%& 62.91\%& 52.42\%\\
\hline
\end{tabular}
\caption{More detailed Transfer Attack results, where the results of the perturbations generated by a model on each branch are shown horizontally, and the results of the model under the perturbations generated by each model are shown vertically.}
\label{tab:bo_effect}
\end{table}

\section{Conclusion}
In this paper, we present BORT, the first adversarial training framework leveraging a multi-branch neural network, which improves network representation through multiple orthogonal branches while keeping inference time and computational cost from increasing, greatly improving model robustness against adversarial attacks.
\bibliographystyle{IEEEtran}
\bibliography{reference}
\end{document}